# Exploiting Particle Swarm Optimization in Multiple Faults Fuzzy Detection

Imtiez Fliss* and Moncef Tagina*

**Abstract**— In this paper an on-line multiple faults detection approach is first of all proposed. For efficiency, an optimal design of membership functions is required. Thus, the proposed approach is improved using Particle Swarm Optimization (PSO) technique. The inputs of the proposed approaches are residuals representing the numerical evaluation of Analytical Redundancy Relations. These residuals are generated due to the use of bond graph modeling. The results of the fuzzy detection modules are displayed as a colored causal graph. A comparison between the results obtained by using PSO and those given by the use of Genetic Algorithms (GA) is finally made. The experiments focus on a simulation of the three-tank hydraulic system, a benchmark in the diagnosis domain.

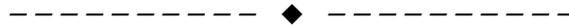

## 1 INTRODUCTION

WITH the rapid progress of science and echnology, the security and reliability of industrial systems become more and more fundamental. Proper and timely fault detection is one of the key technologies that can guarantee system safety and reliability. In fact, development of on-line fault detection systems improves the operational reliability of industrial systems and reduces the operational and maintenance costs and allows detecting faults as quickly as possible.

In this context and with the continuous expansion of industrial applications which are becoming increasingly complex, the problems of multiple faults occur. This means that several faults have simultaneous effects on variables. The number of faulty candidates can grow exponentially and several phenomena of faults compensation can be present. Generally, a single fault changes the performance of overall system. The multiple fault detection is then a challenging problem since a single faulty assumption could lead to incorrect decisions and consequences can be extremely serious in terms of human lives, environmental impact and economic loss. Higher performances and more accurate security necessities have then invoked an ever rising demand to develop on-line multiple faults detection systems.

Fault detection systems consist in deciding whether the system is normally functioning or there are faults. The situation could be more critical if the knowledge about a given system behavior is imperfect. Doubts can arise either about knowledge validity (it is said to be uncertain), or about how to express knowledge clearly (it is vague or imprecise). These two types of imperfections are different but often intimately linked. Noise measurements and modeling approximations are sources of uncertainty in the decision to on-line classify a process state with measurements [8]. It is, then, difficult to distinguish between the effects of an actual fault and those caused by uncertainty and disturbances. Fuzzy logic reasoning is a solution to avoid instable decisions in case of uncertainty and imprecision.

In this context, this paper (which is an extended version of [9]) aims to propose an on-line multiple faults detection approach for dynamic systems. This approach is based on fuzzy reasoning. However, the success of fuzzy fault detection module depends essentially on its parameters such as the fuzzy membership functions.

For efficiency, an optimal design of membership functions is then desired [19]. Therefore, it is useful to use an optimization technique to tune the parameters of the fuzzy membership functions. We choose the Particle Swarm Optimization (PSO) which is an interesting optimization technique. In fact, it is a collective iterative method, with the emphasis on cooperation. It is, generally characterized as a simple concept, easy to implement, and computationally efficient. Unlike the other heuristic techniques, PSO has a flexible and well-balanced mechanism to enhance local exploration abilities [1].

We propose then an on-line multiple faults fuzzy detection approach improved by using the Particle Swarm Optimization technique. The inputs of our detection approaches are residuals. Residuals are consistency indicators that represent the numerical evaluation of the Analytical Redundancy Relations (ARRs).

Analytical Redundancy Relations (ARRs) are symbolic equations representing constraints between different known process variables (parameters, measurements and sources). When a process operates under normal mode conditions, evaluation of the constraints should reveal values within certain small error bounds. Theoretically, these values should be zero in free fault context.

ARRs are obtained from the behavioral model of the system through different procedures of elimination of unknown variables [8]. To get these ARRs, we are based on the bond graph modeling [5] which allows to deal with the enormous amount of equations describing the process behavior and to display explicitly the power exchange between the process components starting from the instrumentation architecture.

*SOIE Laboratory, National School of Computer Sciences, University Campus of Manouba, 2010 La Manouba, University of Tunis, Tunisia. Emails: {Imtiez.Fliss, Moncef. Tagina}@ensi.rnu.tn

The fuzzy detection conclusions are finally displayed through a colored causal graph. In this step, we are inspired from [8], representing the state of system variables through a gradual colored palette from the green nominal state to the red faulty state. This leads to stable decisions and allows gradual information processing that can be easily understood by the human operator and consequently helps him to take proper corrective actions.

To test the performance of the proposed approaches, we rely on a simulation of a benchmark in the diagnosis domain: the three- tank hydraulic system.

This paper is organized as follows: the second and the third sections present respectively the proposed multiple fuzzy faults detection approaches without and with optimization. Section 4 is devoted to expose the contribution of the proposed approaches to literature. Section 5 is concerned in simulation results: we expose the fault detection results we get in case of both proposed approaches with and without PSO. To make evidence of the effectiveness of using PSO, we compare, in section 6, the results given using PSO to those got by Genetic Algorithms (GA). Finally, some concluding remarks are made.

## 2 THE PROPOSED MULTIPLE FAULTS FUZZY DETECTION APPROACH

Fault detection is crucial in the process of system supervision. It consists in deciding if the physical process is faulty or without considering any disturbance. The purpose of detection is then to establish a rule of decision that can detect the earliest possible the passage of a normal operating condition, called Hypothesis H0, to a state of abnormal behavior, where there are failures, noted Hypothesis H1, corresponding to an unpredictable change of some parameters related to the process.
The detection consists of two main phases: the residual generation and evaluation.

### 2.1 Residual generation

This step consists in calculating the residuals which are consistency indicators between recorded measures observations and the model behavior according to [10]. Recalling that residuals are numerical values of ARRs, it is then important to choose carefully the way to get the more accurate ARRs to ensure the best precision and reliability of fault detectors. Thus, we focus on bond graph modeling [5] which is a powerful multidisciplinary tool for modeling, analysis and ARRs generation.
For the generation of Analytic Redundancy Relations directly from the bond graph model, we rely on the procedure described in [16-17] exploiting the constituent relations of set of junctions of the model. Constituent relations are function of known variable (Se, Sf, De and Df) or unknown (the effort and the flow in link with the considered junction) and are deduced by the covering of causal path. The causal path is an alternation of links and basic elements (R, C or I). According to the causality, the variable crossing is the effort or the flow.

The procedure of Analytic Redundancy Relations genera-

tion is expressed as follows:
1. From variables to supervise, chose a type of junction,
2. Choose a junction of this type,
3. Write the constituent relation and express the unknown variables according to the known variables by covering the possible causal paths,
4. Pass to the following junction and repeat the step 3, until the obtaining of sufficiently of Analytic Redundancy Relations (until the obtaining of different signature for the different variable or depletion of junctions).

### 2.2 Residual evaluation (decision)

This decision-making process consists in evaluating residuals. To avoid instable decisions in case of uncertainties and disturbances, several techniques have been used. We consider in this paper, the fuzzy logic technique. In fact, it is an efficient tool for the conversion of uncertain and inaccurate information. It can take into account the uncertainties by the gradual nature of belonging to a fuzzy set.
It is considered as the best framework in which we can handle uncertainties and which also allows the treatment of some incompleteness. Our work consists in proposing a fuzzy approach for on-line multiple faults detection.

### 2.3 Fuzzy residual evaluation for the detection of multiple faults

Fuzzy fault detection consists in interpreting the residuals (which are the detector inputs) by generating a value of belonging to the class AL (alarm) between 0 and 1 that allows to decide whether the measurement is normal or not. The gradual evolution of this variable from 0 to 1 represents the evolution of the variable to an abnormal state.
The fuzzy logic detection approach involves, as shown in figure 1, three successive processes, namely: fuzzification, fuzzy inference, and defuzzification. It leads finally to decide whether the physical process is faulty or not.

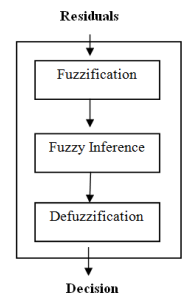

Fig.1. Fuzzy logic detection approach

### 2.3.1 Fuzzification

Fuzzification module transforms the crisp inputs into fuzzy values. In fact, actual input values are mapped into fuzzy membership functions and each input's grade of membership in each membership function is evaluated. Then, these values are processed in the fuzzy domain by inference system, which, is based on the rule base provided by the Knowledge Base (KB).
The inputs of our fuzzy logic detection module are residuals representing the numerical evaluation of the Ana-lytical Redundancy Relations (ARRs).

The Analytical Redundancy Relations (ARR) are symbolic equations which are only function of known variables. They can provide information on the consistency of the model with observations of the system. Therefore they are equal to zeros in normal functioning mode, different of zero otherwise (one or several components of system is or are faulty).

The linguistic set {NB, N, Z, P, PB} meaning negative big, negative, zero, positive and positive big respectively describes the inputs (ARRi) as shown in figure 2.

The supports of the membership functions of the inputs are as the following:

$$\mu_{NB} = [-\beta, -\beta, -ai4, -ai3]$$
$$\mu_N = [-ai4, -ai3, -ai2, -ai1]$$
$$\mu_Z = [-ai2, -ai1, ai1, ai2]$$
$$\mu_P = [ai1, ai2, ai3, ai4]$$
$$\mu_{PB} = [ai3, ai4, \beta, \beta]$$

With $\beta$ a domain characteristic variable which is obtained from experts' knowledge.

Symmetrical trapezoidal membership functions are used for the input fuzzy partitions (see figure 2). Symmetry is justified here by the fact that in the absence of a fault, residuals are zero, and positive or negative faults have the same importance [8]. The symmetry also leads to a simple parameterization of the residual value fuzzy partition with only four parameters: ai1, ai2, ai3 and ai4 (instead of eight) corresponding to the trapezoid boundaries.

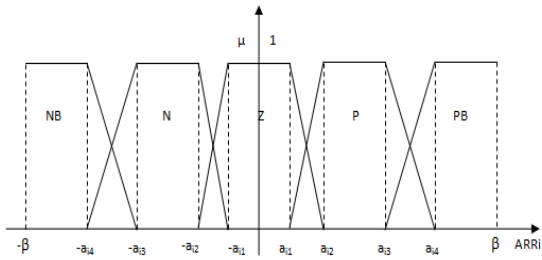

Fig.2. Input Fuzzy partitions

Where:
ai1: maximum value of ARRi in fault-free case.
ai2: minimum value of ARRi in fault case.
ai3: maximum value of ARRi in fault case in limited interval.
ai4: maximum value of ARRi in fault case.
Note that ai1, ai2, ai3 and ai4 are absolute values; they can only be positive.

### 2.3.2 Fuzzy inference system

The second step in the fuzzy logic process involves the inference of the residual values. The fuzzy inference system (FIS) is the responsible for drawing conclusions from the knowledge-based fuzzy rule set of IF-THEN linguistic statements. In case of faults detector, the fuzzy inference system infers variables' states from a set of residuals (ARRs). In fact, the set of fuzzy inputs (corresponding to the N Analytical Redundancy Relations) with their respective membership functions form the premise part of a fuzzy logic analysis. A fuzzy rule set (linguistic if-then statements) is then used to form "judgment" on the fuzzy inputs derived from the N ARRs to get the system variables' states.

In this paper, the MIN-MAX inference method is used. It consists in using the operator min for AND and the operator max for OR.

In this step, we enhance on defining all necessary fuzzy rules in order to detect all single and multiple faults which can affect variables considering especially the cases of faults compensation. All fuzzy rules are therefore determined according to the system functioning analysis.

### 2.3.3 Defuzzification

After determining the relationship between residuals and system variable states and creating the fuzzy connections between the two, the final step is to defuzzify these sets.

Defuzzification is then the process of converting the fuzzy information into crisp values.

The outputs of our fuzzy detection module are the variables' states. An output can be OK or AL. Figure 3 shows the different outputs' membership functions.

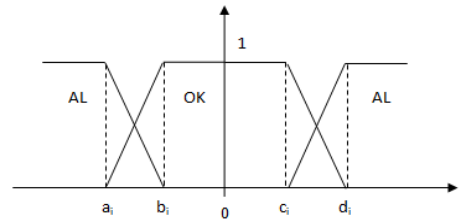

Fig.3. Output Fuzzy Partitions

Where:
ai: minimum negative value of the system variable i in fault case.
bi: minimum negative value of the system variable i in fault-free case.
ci: maximum positive value of the system variable i in fault-free case.
di: minimum positive value of the system variable i in fault case.

The system variable states are defuzzified into an index of a color map from green (normal state) to red (faulty state) and are displayed through a causal graph inspiring from [8].

The use of color code allows operators to analyze a large amount of qualitative information [8].

The layout of the fuzzy logic detector we propose is then shown in figure 4.

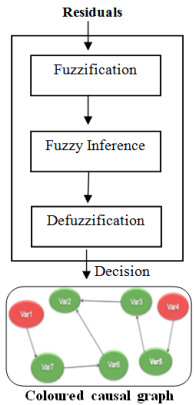

Fig.4. Proposed multiple faults fuzzy detector

The proposed fuzzy logic detector is convenient to identify on-line multiple faults. However, in practical cases, the performance of fuzzy detection is closely linked to the fuzzy membership functions. For efficiency, an optimal design of membership functions is desired [19].

Therefore, we choose to use an optimization technique to adjust the parameters of the fuzzy membership functions of the proposed detector.

## 3 THE PROPOSED MULTIPLE FAULTS FUZZY DETECTION APPROACH WITH OPTIMIZATION

In order to get more efficacy, we point to improve the previously presented multiple faults fuzzy detector by optimally tuning the parameters of inputs and outputs fuzzy partitions. In general, two major categories of optimization algorithms can be distinguished: deterministic and stochastic algorithms [14].

In our work, we consider stochastic methods (also called metaheuristics) which provide a local optimum which may, under certain conditions, be global and are generally characterized by an ease of programming.

In this category of methods, we select the Particle Swarm Optimization (PSO) technique due to its ability to provide solutions efficiently, requiring only minimal implementation effort [14]. Indeed, this metaheuristic differs from other evolutionary methods (typically genetic algorithms) on two essential points: it focuses on cooperation rather than competition and there is no selection (at least in the basic versions).

### 3.1 Overview of Particle Swarm Optimization (PSO)

Particle Swarm Optimization PSO [13-15] is one of the artificial life or multiple agents' type techniques. It is a population-based optimization technique first introduced by Kennedy and Eberhart. It uses a principle of social behavior of a group. A group can achieve the objective effectively by using the common information of every agent, and the information owned by the agent itself.

The motivation for the development of the PSO was based on a simple simulation of animal social behaviors such as bird flocking or fish schooling. Some of the attractive features of PSO include the ease of implementation and the fact that no gradient information is required [3].

The population in PSO is called a swarm and the individuals, referred to as particles, are candidate solutions to the optimization problem at hand in the multidimensional search space (D dimensional).

Each dimension of this space represents a parameter of the problem to be optimized. The position (xi) and velocity (vi) of all particles are usually updated synchronously in each iteration of the algorithm.

A particle adjusts its velocity according to its own flight experience and the flight experience of other particles in the swarm in such a way that it accelerates towards positions that have had high objective (fitness) values in previous iterations.

The algorithm does not require the objective function to be differentiable and continuous and it can be applied to nonlinear and non-continuous cases to solve a wide array of different optimization problems.

There are two kinds of position towards which a particle is accelerated in common use. The first one, a particle's best position achieved up to the current iteration, is called Pbesti. The other is the global best position obtained so far by all particles, called Gbest [18].

### 3.2 Choice of PSO version

There have been several versions of Particle Swarm Optimization. We choose in this paper one of the basic versions as it focuses on cooperation rather than cooperation and is characterized by no selection.

In this context, we choose the PSO version with constriction factor for its speed of convergence [3].

In this case, the modified velocity (vi) and position (xi) of each particle can be calculated using the current velocity, Pbesti and Gbest as shown in the following formulas:

$$v_i^{t+1} = K * (v_i^t + c_1 * rand) * (Pbesti - x_i^{t+1}) + c_2 * rand) * (Gbest - x_i^{t+1})) \quad (1)$$

$$x_i^{t+1} = x_i^t + v_i \quad (2)$$

Where
K: the constriction factor,

$$K = \frac{2}{|2 - c - \sqrt{c^2 - 4c}|} \quad (3)$$

With c=c1+c2 and c>4 [4],

$x_i^t$ : Current position of the ith particle at the t [th] iteration,

$v_i^t$ : Velocity of the particle n° i at the t[th] iteration,

Pbesti: Best previous position of the i[th] particle,

Gbest: Best particle position among all particles in the population,

Rand (): Random number between 0 and 1,

c1: self confidence factor,

c2: swarm confidence factor,

c1 and c2 are positive constants empirically determined.

## 3.3 Fuzzy detection approach using PSO

In this approach, we optimally adjust the inputs' parameters: ai1, ai2, ai3 and ai4 and the output's membership functions parameters aj, bj, cj and dj using PSO with $i \in [\![1..N]\!]$, N is the total number of residuals, $j \in [\![1..P]\!]$ and P is the total number of system variables.

### 3.3.1 PSO Initialization

For parameters initialization of PSO algorithm, we propose the following values:

• Initial swarm population is composed of 30 particles. This choice is based on the study done in [2]. This population size is small enough to be efficient, yet large enough to produce reliable results.
• Initial positions and velocities of particles are generated by using random values.
• The initial fitness value of each particle is initialized to a big positive constant.
• The local best of each particle Pbesti is set to its current fitness value.
• The global best value of the position Gbest is set to the best value of all Pbesti.

### 3.3.2 Fitness function

The fitness function or the objective function is the performance index of a population. The fitness value is bigger, and the performance is better. For the present problem the aim is to have the proper fault decision. Thus, the used fitness function fixes the rate the fault detection error which ought to be minimized.

### 3.3.3 PSO Algorithm

The evolution procedure of PSO Algorithm can be summed up in figure 5 as following:

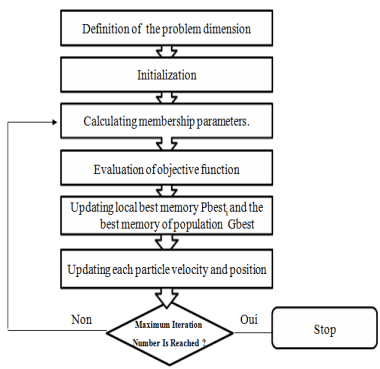

Fig.5. Evolution of PSO algorithm

First of all, the problem dimension and the parameters initialization are defined. Then, the input and output membership parameters are calculated. After evaluating the objective function of solutions, the values of each particle velocity and position are updated using formulas

(1) and (2). If the maximum iteration number is reached, the final solutions are given. Otherwise, we loop to calculating membership parameters and so on,

### 3.3.4 Fuzzy fault detection approach using PSO

In this approach summed up in figure 6, each membership function input and output parameters are optimized according to PSO algorithm. Then, the multiple faults fuzzy detector is applied. Its three successive processes, namely: fuzzification, fuzzy inference, and defuzzification previously presented are used to finally display the resulted decisions as a colored causal graph.

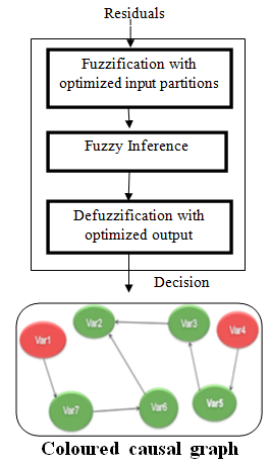

Fig.6. Proposed multiple faults fuzzy detector using optimization

# 4 SIMULATIONS

## 4.1 Considered example

To illustrate the applicability of the presented methods the three-tank hydraulic system is taken as an example.

### 4.1.1 Process Description

The process, shown in figure 7, consists of three cylindrical tanks (Tank1, Tank2 and Tank3) having equivalent cross section A. Tanks communicate through two feeding valves (V4 and V5) with the same cross section Sn. An outflow valve V6 located at Tank 2 has a circular cross section Snout.
Pumps 1, 2 and 3 are controlled by the valves V1, V2 and V3.
The flow liquid rate from tank i to tank j is given by:

$$Q_{ij} = a_z * S * \text{sgn}(h_i - h_j) * \sqrt{2 * g * |h_i - h_j|} \quad (4)$$

Where:
- hi (measured in meters) is the liquid level of tank i for i=1, 2, 3, respectively.
- az the outflow coefficient.
- S is the sectional area of the connecting valve.
- g the gravitational constant.

Valves V1, V2, V3, V4, V5 and V6 are functioning in binary mode on or off. Fluid levels in the three tanks can be low, medium or high. When V4 opens, it automatically closes itself after pouring enough liquid so that the liquid level in the Tank3 increases. Similarly, when V5 opens, it automatically shuts itself down after pouring enough liquid so that the liquid level in the Tank3 increases. The global purpose of the three- tank system is to keep a steady fluid level in the Tank 3, the one in the middle.

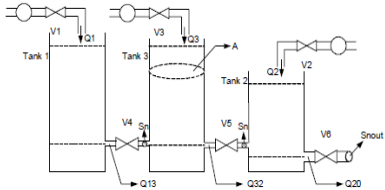

Fig.7. The three-tank hydraulic system [6]

Msf1 and Msf2, corresponding to volume flows applied to the system, are the two inputs of the process. We put five sensors: effort sensors De1, De2 and De3 to measure pressure of C1:tank1, C2:tank3 and C3:tank2 and flow sensors Df1and Df2 measuring flow level of the valve 1 and valve 2 (see figure 8).

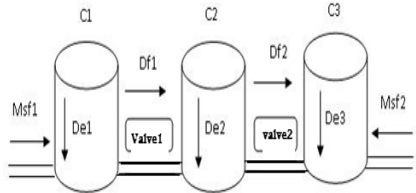

Fig.8. The three-tank system

### 4.1.2 Bond Graph Modeling
The bond graph model of the three- tank system is giving in figure 9.

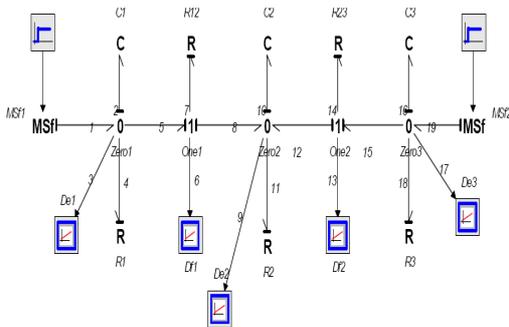

Fig.9. Bond Graph model of the three-tank system

The tanks are modeled as capacitances that hold fluid, and the valves are modeled as resistances to flow.

0− and 1− junctions represent the common effort (i.e., pressure) and common flow (i.e., flow rate) points in the system, respectively. Each junction in the reference model represents a constitutive equation for efforts or flows. Measurement points, shown as Dei and Dfi components, are respectively effort detectors and flow detectors not consuming power (supposed ideal) and connected to junctions.

Thanks to structural behavioral and causal properties [11] of Bond Graph, the causal graph of the three- tank process can be generated as given in figure 10.

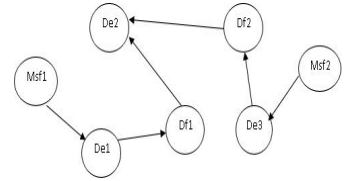

Fig.10. Influence Graph of the three- tank system

In this case, the graph nodes represent the system variables; the directed arcs symbolize the normal relations among them (for instance Msf1->De1 means that modifications of Msf1 necessary cause changes of De1).

### 4.1.3 ARR determination
For the generation of Analytic Redundancy Relations directly from the bond graph model, we are based on the bond graph model given in figure 9 and the procedure described in [16-17] and previously presented. We obtain the following Analytical Redundancy Relations:

ARR1:

$$\frac{1}{sC_1}MSf_1 - (1 + \frac{1}{R_1C_1s})De_1 - \frac{1}{C_1s}Df_1 = 0 \qquad (5)$$

ARR2:

$$\frac{1}{sC_2}Df_1 - (1 + \frac{1}{R_2C_2s})De_2 - \frac{1}{C_2s}Df_2 = 0 \qquad (6)$$

ARR3:

$$\frac{1}{C_3s}MSf_2 - \frac{1}{C_3s}Df_2 - (1 + \frac{1}{R_3C_3s})De_3 = 0 \qquad (7)$$

ARR4:

$$\frac{De_3 - De_2}{R_{23}} - Df_2 = 0 \qquad (8)$$

ARR5:

$$\frac{De_1 - De_2}{R_{12}} - Df_1 = 0 \qquad (9)$$

The bond graph model is then converted into synopsis diagram simulated within Simulink/Matlab environment.

## 4.2 Application choices

### 4.2.1 Disturbance modeling
Disturbance is an important criterion for testing the performance of detection and isolation techniques. Through our tests, we disturb the parameters R and C with a Gaussian white noise.

### 4.2.2 Fault modeling
There are two ways to model faults. The first one assumes that faults are modeled by parasite signal inputs. This is known as additive faults. The second way assumes that faults modify the system model. This change usually takes the form of a modification of model parameters, more rarely, a modification of the model structure. We referred to as multiplicative faults [11].

We consider in our work additional faults modeled as additive signals added to the three- tank system variables.

## 4.3 Simulation results

To test the performance of the presented approaches, we use 50 faults scenarios. At each test, we inject single or multiple faults, and make a note of detection results.

In free-fault context, it is checked that all residuals describing Analytical Redundancy Relations are null (see figure 11).

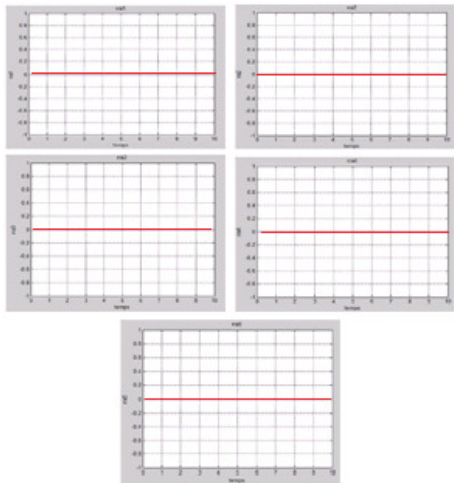

Fig.11. Free- fault context

All over the made tests, we checked if proper decision is finally given. If so, we note the fault detection delay.

In the following sections, we first expose the results given in case of using our multiple faults fuzzy detection approach without optimization then, we introduce those given in case of using optimization.

### 4.3.1 Multiple Faults Fuzzy detection without optimization

Using the proposed fuzzy detection approach without optimization, we get proper decision in 39/50 cases. For the rests, the results are either bad detection (i.e. the system is announced normally functioning while there are actually faults) or missed detection (not all faults are identified) cases.

For instance, the multiple faults fuzzy approach without optimization gives a bad detection in case of faults affecting De2 and Msf2. In this case, the proposed approach does not detect any faults. The corresponding result is given in figure 12.

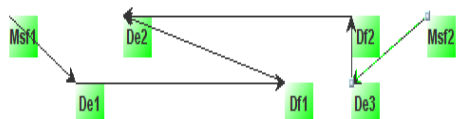

Fig.12. Result of injecting De2 and Msf2 using the fault detection module without optimization

It gives also a missed detection in case of injecting faults on {De1, Df1 and Msf2}, detecting only faults on Msf2. The colored graph displaying the result of this case is given in figure 13.

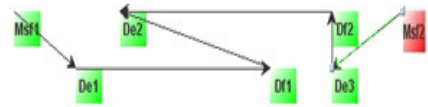

Fig.13. Result of injecting {De1, Df1 and Msf2} using the fault detection module without optimization

We also notice important faults detection delays. For example, a delay of 10 seconds in fault detection is noticed in case of injecting Msf1, Msf2 and De2 at 30s. In case of injecting Msf2 and Df2 at 50s, the delay reaches 17 seconds.

### 4.3.2 Multiple Faults Fuzzy detection using PSO

In case of using the on-line multiple fault fuzzy detection with PSO, it is necessary first of all to tune the parameters of the fuzzy fault detector.

#### 4.3.2.1 The optimal membership functions

The membership functions' parameters are [ai1, ai2, ai3, ai4] for each input (in this case we have five inputs {ARR1, ARR2, ARR3, ARR4 and ARR5}) and [aj, bj, cj and dj] for each output {Msf1, Msf2, De1, De2, De3, Df1 and Df2}.

Table 1 lists the parameters of the PSO algorithm used in our work. We base our choice of the population size, c1 and c2 on the study done in [2].

TABLE 1
PARAMETERS OF PSO ALGORITHM

| Population Size | 30 |
|---|---|
| Number of Iterations | 10000 |
| c1 | 2.8 |
| c2 | 1.3 |
| ai1 with i∈ [ \|1,5\| ] | [0, 1] |
| ai2 with i∈ [ \|1,5\| ] | [0, 1.5] |
| ai3 with i∈ [ \|1,5\| ] | [1, 5] |
| ai4 with i∈ [ \|1,5\| ] | [1.5, 5] |
| aj with j∈ [ \|1,7\| ] | [-5, 0] |
| bj with j∈ [ \|1,7\| ] | [-0.5, 0] |
| cj with j∈ [ \|1,7\| ] | [0, 5] |
| dj with j∈ [ \|1,7\| ] | [0.5, 5] |

With {ai1, ai2, ai3, ai4} the parameters of the input n° i; {i∈ [ \|1, 5\| ]} and {aj, bj, cj, dj} the parameters of the j$^{th}$ output; {j∈ [ \|1, 7\| ]}.

After the evolution process of PSO, we obtain as a result:

TABLE 2
RESULTS OF THE APPLICATION OF PSO

| a11 | 0.146 | a51 | 0.225 | a4 | -0.701 |
|---|---|---|---|---|---|
| a12 | 0.973 | a52 | 1.05 | b4 | -0.304 |
| a13 | 1.57 | a53 | 1.468 | c4 | 0.304 |
| a14 | 1.73 | a54 | 1.87 | d4 | 0.675 |
| a21 | 0.516 | a1 | -0.754 | a5 | -0.674 |
| a22 | 1.343 | b1 | -0.304 | b5 | -0.357 |
| a23 | 1.944 | c1 | 0.304 | c5 | 0.251 |
| a24 | 2.24 | d1 | 0.6746 | d5 | 0.542 |
| a31 | 0.225 | a2 | -0.622 | a6 | -0.463 |
| a32 | 1.057 | b2 | -0.172 | b6 | -0.146 |
| a33 | 1.657 | c2 | 0.383 | c6 | 0.462 |
| a34 | 1.957 | d2 | 0.674 | d6 | 0.753 |
| a41 | 0.49 | a3 | -0.621 | a7 | -0.621 |
| a42 | 1.32 | b3 | -0.357 | b7 | -0.3305 |
| a43 | 1.92 | c3 | 0.251 | c7 | 0.2275 |
| a44 | 2.42 | d3 | 0.595 | d7 | 0.595 |

The optimal membership functions consequently obtained are shown in figure 14.

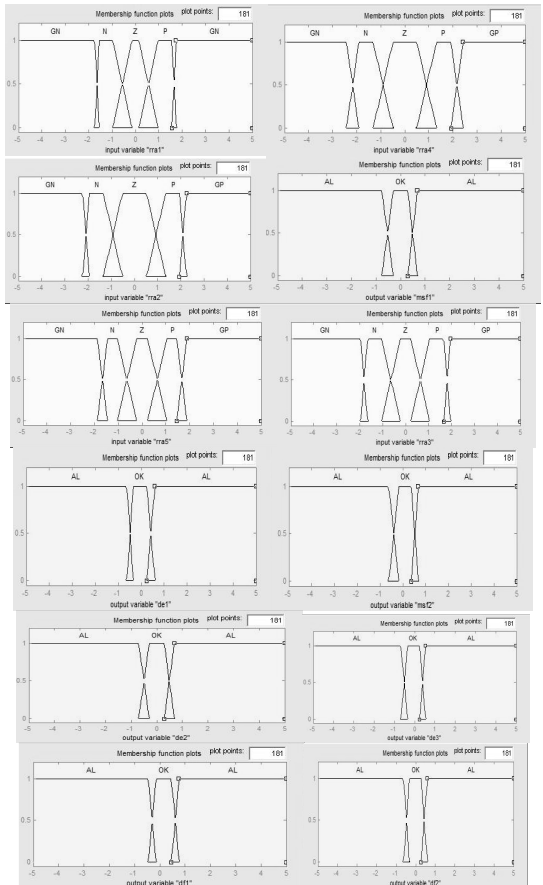

Fig.14. The optimal membership functions

The first five pictures in figure 14 show the optimized fuzzy input partitions and the last seven pictures are the optimized fuzzy output partitions. Once the parameters of membership functions are adjusted, we use fuzzy rea-soning (to determine the status of the system: normal or faulty displaying the results as a colored causal graph representing the state of the different variables through a gradual palette of colors from the green nominal state to the red faulty state).

**4.3.2.2 Multiple Faults Fuzzy detection with PSO**
The detection results are satisfactory in this case. We get, indeed, the proper decision in 49/50 of cases. For instance figure 15 and figure16 show faults detection results of the multiple faults fuzzy detector using PSO when faults affect {De2 and Msf2} and {Msf1, Msf2 and De2} respectively.

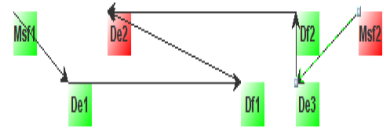

Fig.15. Result of injecting De2 and Msf2 using the fault detection module with PSO

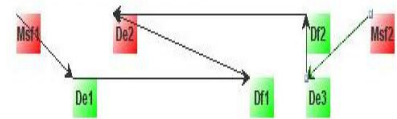

Fig.16. Result of injecting Msf1, Msf2 and De2 using the fault detection module with PSO

The delay in fault detection is also reduced. In case of injecting faults to Msf1, Msf2 and De2, we notice a proper decision at time without any delay.
The results we get prove evidence of the effectiveness of the use of PSO to improve fault detection.

**4.4 Discussion**
Comparison between the results obtained using the multiple fault detection approaches without and with PSO in the case of the three- tank system is resumed in figure 17 and figure 18.

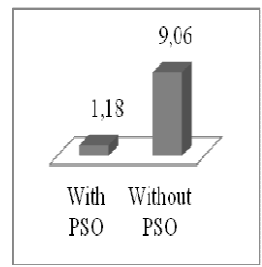

Fig.17. Comparison of average delay in detection in case of using the proposed fuzzy detector with and without PSO

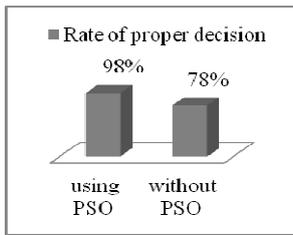

Fig.18. Comparison of rate of proper decision in case of using the proposed fuzzy detector with and without PSO

Results prove that both proposed multiple faults fuzzy detection approaches give good rate of proper decision (more than 75% of cases). However, we notice that the use of PSO in our on-line multiple faults Fuzzy detection module improves the results and gives the proper decisions in majority of cases with a small delay. It is also clear that the use of PSO presents an average improvement of 20% of proper detections giving in the case of the use of our approach without PSO and a reducing of fault delay of proximately 8 s.

The application of the PSO algorithm in our case shows that this technique presents a great promise for detection process. In fact, the use of PSO in our study case improved the performance of the proposed fuzzy fault detection module, decreases the possibility of bad and missed detections and reduced the detection delay.

To make evidence of the efficiency of using PSO, it would be interesting to compare its performance in case of multiple faults detection with an evolutionary technique that has attracted considerable attention: the Genetic Algorithms (GA).

## 5 MULTIPLE FAULTS FUZZY DETECTION WITH GA

### 5.1 Overview of Genetic Algorithms

The GA is an optimization routine based on the principles of Darwinian Theory and natural genetics. It has primarily been utilized as an off-line technique for performing a directed search for the optimal solution to a problem. The GA performs a parallel, directed, random search for the fittest element of a population within a search space. The population simply consists of strings of numbers, called chromosomes that hold possible solutions of a problem [7].

Like PSO, the GA begins its search from a randomly generated population of designs that evolve over successive generations (iterations), eliminating the need for a user-supplied starting point. To perform its optimization-like process, the GA employs three operators called selection, crossover and mutation to propagate its population from one generation to another [12].

By creating successive generations which continue to evolve, the GA will tend to search for a global optimal solution. The key to the search is the fitness evaluation. The fitness of each of the members of the population is calculated using a fitness function that characterizes how well each particular member solves the given problem. Parents for the next generation are selected based on the fitness value of the strings. That is, strings that have a higher fitness value are more likely to be selected as parents, and, thus, are more likely to survive to the next generation.

### 5.2 Fuzzy fault detection approach using GA

In this approach, we also intend to optimally tune the inputs' parameters: ai1, ai2, ai3 and ai4 and the output's membership functions parameters aj, bj, cj and dj using GA with $i \in [1..N]$, N is the total number of residuals, $j \in [1..P]$ and P is the total number of system variables.

#### 5.2.1 Genome representation
Each individual or chromosome (genome) represents a possible solution. Every chromosome X is made up of a sequence of genes. Within our work, each chromosome contains the parameters {ai1, ai2, ai3, ai4} of each input i and {aj, bj, cj, dj} parameters of each output j.

#### 5.2.2 GA initialization
To start genetic algorithm, an initial population is needed. Population initialization is the procedure where the first generation of the population is determined within the search space. Within our study, the initial population is randomly generated. It is composed of 30 individuals.

#### 5.2.3 Fitness function
The used fitness function is the same one used in case of Particle Swarm Optimization. Thus, it fixes the rate the fault detection error which ought to be minimized.

#### 5.2.4 GA selection function
Selection is the procedure of selecting individuals from the population to form a pool of parents that will be used to produce offspring through recombination.

To produce successive generations, selection of individuals plays a significant role. The selection function determines which of the individuals move on to the next generation. In the present work, ranked replacement was considered.

#### 5.2.5 GA operators
In GA, there are generally two types of operators: crossover and mutation. These operators are used to produce new solutions based on existing solutions in the population.

- Crossover or recombination is the procedure of recombining the information carried by two individuals to produce new offspring. This is in direct analogy to the biological reproduction, where DNA sequences of parents are mixed to produce offspring DNA sequences that combine their genetic information [14]. Crossover takes two individuals to be parents and produces two new individuals. An intermediate crossover operator is used in our study.

- Mutation, on the other hand, alters one individual to produce a single new solution. According to [14], mutation is a fundamental biological operation. It enables organisms to change one or more biological properties radi-cally, in order to fit an environmental change or continue their evolution by producing offspring with higher chances of survival. In nature, mutation constitutes an abrupt change in the genotype of an organism, and it can be either inherited by parents to children or acquired by an organism itself. A DNA mutation may result in the modification of small part(s) of the DNA sequence or rather big sections of a chromosome. Its effect on the organism depends heavily on the mutated genes. While mutations to less significant genes have small positive or negative effects, there are mutations that trigger radical changes in the behavior of several genes. These mutations alternate genes that control the activation of other genes. Thus, they have a crucial impact that can possibly affect the whole structure of the organism.

Within our work, uniform mutation operator is used.

### 5.2.6 GA Algorithm

The following outline summarizes how the genetic algorithm we use works:

1. The algorithm begins by creating a random initial population.
2. The algorithm then creates a sequence of new populations. At each step, the algorithm uses the individuals in the current generation to create the next population. To create the new population, the algorithm performs the following steps:
   a) Scores each member of the current population by computing its fitness value.
   b) Scales the raw fitness scores to convert them into a more usable range of values.
   c) Selects members, called parents, based on their fitness.
   d) Some of the individuals in the current population that have lower fitness are chosen as elite. These elite individuals are passed to the next population.
   e) Produces children from the parents. Children are produced either by making random changes to a single parent mutation or by combining the vector entries of a pair of parents crossover.
   f) Replaces the current population with the children to form the next generation.
3. The algorithm stops when one of the following stopping criteria is met:
   a) The maximum number of generations is reaches (in our study we use 100 as a maximum number).
   b) There is no improvement in the best fitness value for 50 generations.

### 5.3 Parameters optimization: simulation

The first step in our work is to determine the membership functions' parameters which are [a1, a2, a3, a4] for each input and [a, b, c and d] for each output.

Table 3 lists the parameters of GA used in our work.

TABLE 3

PARAMETERS OF GA

| Population Size | 30 |
|---|---|
| Maximum number of generations | 100 |

| Genome representation | X |
|---|---|
| Mutation function | Uniform mutation |
| Crossover function | Intermediate |
| Scaling function | Rank |

With the genome representation X= [a11, a12, a13, a14, a21, a22, a23, a24, a31, a32, a33, a34, a41, a42, a43, a44, a51, a52, a53, a54, a1, b1, c1, d1, a2, b2, c2, d2, a3, b3, c3, d3, a4, b4, c4, d4, a5, b5, c5, d5, a6, b6, c6, d6, a7, b7, c7, d7].

After the evolution process of GA described in (5.2.6) and using the parameters given in table 3, we obtain as a result:

TABLE 4

SIMULATION RESULTS

| a11 | 0.106 | a51 | 0.325 | a4 | -0.721 |
|---|---|---|---|---|---|
| a12 | 0.93 | a52 | 1.214 | b4 | -0.2021 |
| a13 | 1.07 | a53 | 1.68 | c4 | 0.398 |
| a14 | 1.73 | a54 | 1.87 | d4 | 0.668 |
| a21 | 0.436 | a1 | -0.754 | a5 | -0.654 |
| a22 | 1.403 | b1 | -0.42 | b5 | -0.326 |
| a23 | 1.84 | c1 | 0.2019 | c5 | 0.281 |
| a24 | 2.42 | d1 | 0.7146 | d5 | 0.526 |
| a31 | 0.325 | a2 | -0.628 | a6 | -0.429 |
| a32 | 1.07 | b2 | -0.189 | b6 | -0.257 |
| a33 | 1.67 | c2 | 0.283 | c6 | 0.392 |
| a34 | 1.786 | d2 | 0.474 | d6 | 0.543 |
| a41 | 0.409 | a3 | -0.528 | a7 | -0.671 |
| a42 | 1.29 | b3 | -0.343 | b7 | -0.2105 |
| a43 | 1.928 | c3 | 0.154 | c7 | 0.3275 |
| a44 | 2.53 | d3 | 0.498 | d7 | 0.492 |

### 5.4 Multiple Faults Fuzzy detection using GA

The detection approach using Genetic Algorithms gives a small rate of bad detections 8%. We can say then that the results are acceptable. For instance figure 19 shows faults detection results when faults affect {De2 and Msf2}.

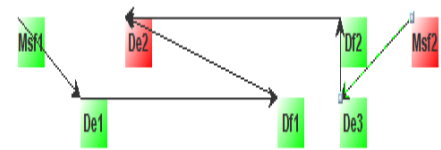

Fig.19. Result of injecting De2 and Msf2 using the fault detection module using GA

In this case, we notice a proper decision with a delay in detection of 2s in case of injecting Msf2 and De2 at 40s.

### 5.5 Discussion

Comparisons between the results obtained using the multiple fault detection approaches with PSO and with GA in the case of the three- tank system are resumed in figures 20 and figure 21.

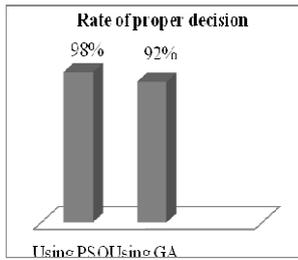

Fig.20. Comparison between genetic algorithms and PSO: Proper decision rate

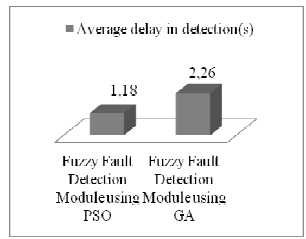

Fig. 21. Comparison between genetic algorithms and PSO: average delay in detection

We notice that both used optimization techniques give good rate of proper decision (more than 90%) and a small average delay in detection (less than 3s) in the case of the three-hydraulic system with the choices we made in tables1 and 3.

According to the simulation results given in our study case, PSO proves its effectiveness giving the proper fault decision in 49/50 cases and reducing the delay in detection. This can be explained by the fact that the best finally solutions (membership functions parameters) are obtained from bad intermediate solutions that were eliminated in the case of GA.

This influences the performance of the multiple faults fuzzy detection particles which is sensitive to its membership functions parameters.

Dealing with the implementation aspects, it is to notice that, PSO is easier to program compared to GA. and there are few parameters to adjust which match with [12].

In fact, PSO does not have genetic operators like crossover and mutation. Particles update themselves with the internal velocity. They also have memory, which is important to the algorithm.

We can then conclude that the use of PSO presents a great improvement and promise of our on-line multiple faults fuzzy detection approach.

## 6 CONTRIBUTION TO THE LITERATURE

The main contribution of our work concerns the detection of multiple faults in dynamic systems. The considered multiple faults occur at the same time in situations characterized by imperfections. Thus, we define two fuzzy approaches which benefits from the pros of the fuzzy reasoning and which results are given as a gradual evolution of belonging to alarm classes and displayed through a colored causal graph giving representative information for decision-making to the operator. In fact, thanks to capability of fuzzy logic to describe vague and imprecise facts, the proposed technique is also well adapted to get the most accurate decision even if residuals are affected by the noise contamination and uncertainty effects [10].

The proposed fuzzy approach is inspired from the work of [8] in presenting the results as a colored causal graph. Whereas the proposition of [8] did not concentrate on the multiple faults detection, our approach focus essentially on the detection of multiple faults that occurs simultaneously and which can have compensation effects. Thus, the proposed fuzzy logic schema is different from the one proposed in Evsukoff et al. work  especially the fuzzy inference process as we  consider especially the case of faults effects compensation ( therefore, all possible fault cases  are studied in the fuzzy inference system according to the analysis of the system functioning information).

The performance of fuzzy detection is closely related to the fuzzy membership functions parameters.  Therefore, we choose to use PSO technique (the Particle Swarm Optimization) to adjust the parameters of the fuzzy membership functions. This technique improves the performance of our fuzzy detection module and consequently the performance of the whole proposed detection technique [10]. There have been several novelties in Particle Swarm Optimization. However, we choose in this paper one of the basic versions as it focuses on cooperation rather than cooperation and is characterized by no selection. The comparison with genetic algorithm optimization technique proves the efficacy of PSO and highlights the choice of the used version of Particle Swarm Optimization.

## 7 CONCLUSION

In this paper, we propose a fuzzy approach for on-line multiple faults detection. In fact, fuzzy systems are convenient and efficient for solution for such a problem.

However, the major drawback of the fuzzy logic systems is insufficient analytical technique design (like parameterization of the membership functions).

Therefore, an improvement of the proposed approach using Particle Swarm Optimization is presented in order to adjust the membership functions' parameters of a fault detection module.

The inputs of the detection modules are Analytical Redundancy Relations obtained from bond graph model.

The conclusions of the fuzzy reasoning detection module are then defuzzified into an index of a color map from green (normal state) to red (faulty state) and given as a colored causal graph.

A simulation of the three-tank system proves that the improved multiple faults fuzzy detection module gives very good results compared to the one without optimization.

Finally a comparison with GA is made, for this application concerning the three tanks hydraulic system, using PSO demonstrates its superiority. In fact, simulations show a good rate for proper detection and a small delay in detection in case of using PSO with regards to using Genetic Algorithms.

It is expected that the achieved results can also be extended to detecting and localizing multiple faults in hybrid systems. In fact, we intend in future works to highlight the potential of using such a diagnosis solution hybrid applications.

# 8 REFERENCES

[1] Abido, M. A., 2002. Optimal Design of Power-System Stabilizers Using Particle Swarm Optimization. IEEE Trans. Energy Conversion, Vol. 17, p. 406- 413.

[2] Carlisle, A., and G. Dozier, 2001. An Off-The-Shelf PSO. Proceedings of the Particle Swarm Optimization Workshop, Indianapolis, Ind, Usa, p. 1–6.

[3] Clerc, M., 1999. The Swarm and the Queen: Towards A Deterministic and Adaptive Particle Swarm Optimization. Proceedings of the Congress of Evolutionary Computation, Washington, Dc, p. 1951–1957.

[4] Clerc, M., and J. Kennedy, 2002. The Particle Swarm: Explosion, Stability and Convergence in a Multi- Dimensional Complex Space. IEEE Transactions on Evolutionary Computation, Vol. 6.

[5] Dauphin-Tanguy, G., 2000. Les Bond Graph. Hermes Sciences Publications.

[6] Dotoli M., Fanti M. P., and Mangini A. M., 2006, "On-line identification of discrete event systems: a case study", in Proceedings of the IEEE International Conference on Automation Science and Engineering (CASE '07), p. 405–410, Shangai, China.

[7] El-Telbany, M.E.-S., 2007. Employing Particle Swarm Optimizer and Genetic Algorithms for Optimal Tuning of PID Controllers: A Comparative Study. ICGST-ACSE Journal, Volume 7, Issue 2, p. 49-53.

[8] Evsukoff, A., S. Gentil and J. Montmain, 2000. Fuzzy Reasoning in Cooperative Supervision Systems. Control Engineering Practice, p. 389- 407.

[9] Fliss I. and Tagina M., Multiple faults fuzzy detection approach improved by Particle Swarm Optimization, The 8th International Conference of Modelling and Simulation - MOSIM'10, Hammamet, Tunisia, May 10-12, 2010.

[10] Fliss I. and Tagina M., "Multiple faults model-based detection and localization in complex systems", Journal of Decision Systems (JDS), vol. 20 Issue 1/2011, pp.7-31.

[11] Gentil, S., 2007. Supervision des procédés complexes, Hermès Science Publication, Paris.

[12] Hassan, R., B. Cohanim and O. De Weck, 2005. A Comparison of Particle Swarm Optimization and the Genetic Algorithm. 46th Aiaa/Asme/Asce/Ahs/Asc Structures, Structural Dynamics & Materials Conference, Austin, Texas.

[13] Kennedy, J., and R.C. Eberhart, 1995. Particle Swarm Optimization. Proceeding of The IEEE International Conference On Neural Networks, Perth, Australia, Vol. 4, p. 1942- 1948.

[14] Parsopoulos K. E and Vrahatis M. N, 2010. Particle Swarm Optimization and Intelligence: Advances and Applications. Information Science Reference (an imprint of IGI Global), United States of America.

[15] Shi, Y., and R.C. Eberhart, 1998. A Modified Particle Swarm Optimiser. IEEE International Conference on Evolutionary Computation, Anchorage, Alaska.

[16] Tagina, M., J.P. Cassar, G. Dauphin-Tanguy and M. Staroswiecki, 1996. The Bond Graph Use for the Design and The Improvement Of Instrumentation Architecture For System Monitoring. Int Mechanical Engineering Congress and Exhibition, Publication Dans Dynamics Systems and Control Division's Proceedings, Atlanta, Georgia, USA.

[17] Tagina, M., J.P. Cassar, G. Dauphin-Tanguy and M. Staroswiecki, 1996. Bond Graph Models For Direct Generation Of Formal Fault Detection Systems. International Journal of Systems Analysis Modelling and Simulation, Vol.23, p. 1- 17.

[18] Yisu, J., J. Knowles, L. Hongmei, L. Yizeng and D.B. Kell, 2008. The Landscape Adaptive Particle Swarm Optimizer. Applied Soft Computing, 8, p.295- 304.

[19] Zhou, Y.S., and L.Y. Lai, 2000. Optimal Design for Fuzzy Controllers by Genetic Algorithms. IEEETrans. On Industry Application, Vol. 36, No. 1, p. 93 - 97.

**Imtiez FLISS** is currently preparing her PhD thesis and teaching at the National School of Computer Sciences (Tunisia). She studied computer sciences at National School of Computer Sciences (Tunisia) where she obtained the Engineering degree in 2008 and the Master degree in 2009. Her research interests include multiple faults detection and localization in continous, discrete and hybrid complex systems and Artificial Intelligence.

**Prof. Moncef Tagina** is a Professor at the National School of Computer Sciences (Tunisia). He went to the Ecole Centrale de Lille (France), where he studied Engineering and obtained his degrees and Master of Automatic in 1992 and PhD thesis in 1995. He teached one year at Ecole Centrale de Lille moving in 1996 to University of Science of Monastir. He obtained his"University Habilitation" from National School of Engineers of Tunis in 2002. His research interests include bond graph modeling, Systems monitoring, robotics and Artificial Intelligence.